\documentclass[preprint,12pt,3p]{elsarticle}
\usepackage{subcaption}
\usepackage{url}

\usepackage{subfiles}

\input{tikzstuff}

\usepackage{booktabs}
\usepackage{multirow}
\usepackage{todonotes}
\usepackage{amssymb}
\usepackage{caption}
\usepackage{lipsum}
\usepackage{amsmath,amsfonts}
\usepackage{xcolor}

\DeclareMathOperator*{\argmin}{arg\,min}

\journal{Medical Image Analysis}

\begin{document}

\begin{frontmatter}

\title{Automatic Liver and Tumor Segmentation of CT and MRI Volumes Using Cascaded Fully Convolutional Neural Networks}

\author[label1]{Patrick Ferdinand Christ\fnref{fn1}}
\author[label1]{Florian Ettlinger\fnref{fn1}}
\author[label1]{Felix Gr\"un}
\author[label1]{Mohamed Ezzeldin A. Elshaer}
\author[label1]{Jana Lipkov\'a}

\author[label1]{Sebastian Schlecht}
\author[label1]{Freba Ahmaddy}
\author[label1]{Sunil Tatavarty}
\author[label1]{Marc Bickel}
\author[label1]{Patrick Bilic}
\author[label1]{Markus Rempfler}
\author[label2]{Felix Hofmann}
\author[label2]{Melvin D'Anastasi}

\author[label2]{Seyed-Ahmad Ahmadi}
\author[label1]{Georgios Kaissis}
\author[label2]{Julian Holch}
\author[label2]{Wieland Sommer}
\author[label1]{Rickmer Braren}
\author[label2]{Volker Heinemann}
\author[label1]{Bjoern Menze}

\address[label1]{Technical University of Munich, Arccistrasse 21, 80333 Munich}
\address[label2]{LMU Hospital Grosshadern, Marchioninistrasse 15, 81377 Munich, Germany}

\fntext[fn1]{Authors contributed equally}
\begin{abstract}
Automatic segmentation of the liver and hepatic lesions is an important step towards deriving quantitative biomarkers for accurate clinical diagnosis and computer-aided decision support systems. This paper presents a method to automatically segment liver and lesions in CT and MRI abdomen images using cascaded fully convolutional neural networks (CFCNs) enabling the segmentation of large-scale medical trials and quantitative image analyses.
We train and cascade two FCNs for the combined segmentation of the liver and its lesions. As a first step, we train an FCN to segment the liver as ROI input for a second FCN. The second FCN solely segments lesions within the predicted liver ROIs of step 1.
CFCN models were trained on an abdominal CT dataset comprising 100 hepatic tumor volumes. Validation results on further datasets show that CFCN-based semantic liver and lesion segmentation achieves Dice scores over $94\%$ for the liver with computation times below 100s per volume. We further experimentally demonstrate the robustness of the proposed method on 38 MRI liver tumor volumes and the public 3DIRCAD dataset.

\end{abstract}


\end{frontmatter}



\newcommand{\labelspace}{\mathcal{\mathcal{L}}}     
\newcommand{\x}{\mathbf{x}} 
\newcommand{\vertices}{\mathcal{V}} 
\newcommand{\edges}{\mathcal{E}}    
\newcommand{\graph}{\mathcal{G}}    
\newcommand{\cp}[2]{P\left( #1 \vert #2 \right)}    

\section{Introduction}

\subsection{Motivation}
Anomalies in the shape and texture of the liver and visible lesions in computed tomography (CT) and magnetic resonance images  (MRI) images are important biomarkers for initial disease diagnosis and progression in both primary and secondary hepatic tumor disease \cite{Heimann}.

Primary tumors such as breast, colon and pancreas cancer often spread metastases to the liver during the course of disease. Therefore, the liver and its lesions are routinely analyzed in primary tumor staging. In addition, the liver is also a site of primary tumor disease such as  Hepatocellular carcinoma (HCC). Hepatocellular carcinoma (HCC) presents the sixth-most common cancer and the third-most common cause of cancer-related deaths worldwide \cite{ferlay2010estimates}. HCC comprises a genetically and molecularly highly heterogeneous group of cancers that commonly arise in a chronically damaged liver. Importantly, HCC subtypes differ significantly in clinical outcome. The stepwise transformation to HCC is accompanied by major changes in tissue architecture including an increase in cellularity and a switch in vascular supply (i.e. arterialization). These quantifiable changes in tissue architecture provide the basis for the non-invasive detection of HCC in imaging \cite{european2012easl}, but also lead to highly variable structures and shapes.

\begin{figure}

\minipage{0.32\textwidth}
        \includegraphics[width=\textwidth]{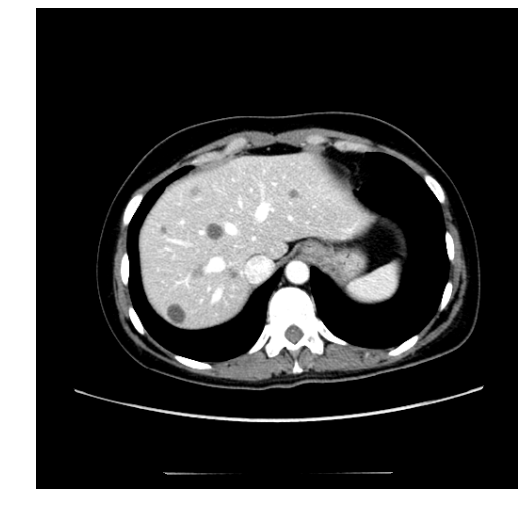}
\endminipage\hfill
\minipage{0.32\textwidth}
        \includegraphics[width=\textwidth]{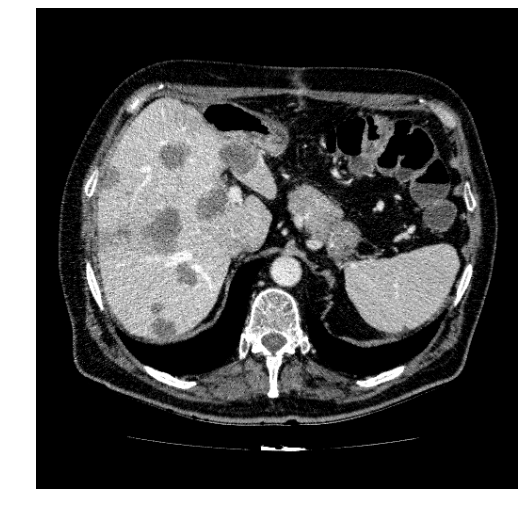}
\endminipage\hfill
\minipage{0.32\textwidth}
        \includegraphics[width=\textwidth]{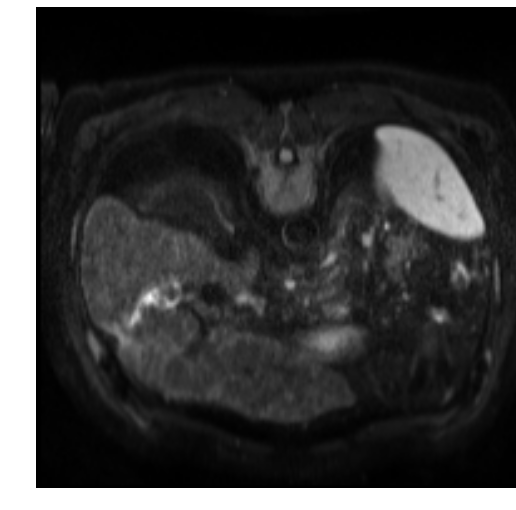}
\endminipage\hfill
\minipage[c]{0.32\textwidth}
\centering
        \includegraphics[width=\textwidth]{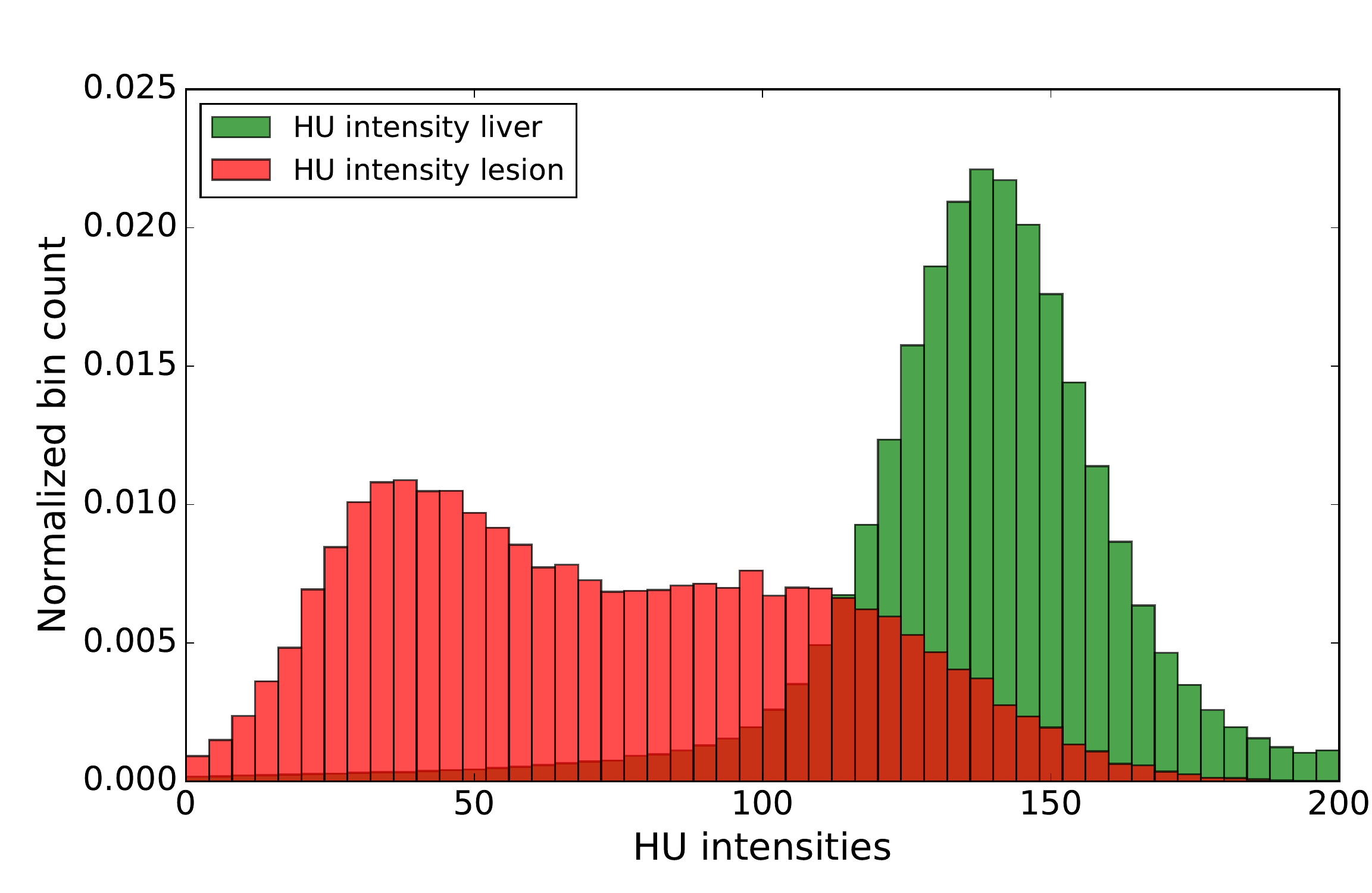}

\endminipage\hfill
\minipage[c]{0.32\textwidth}\
\centering
        \includegraphics[width=\textwidth]{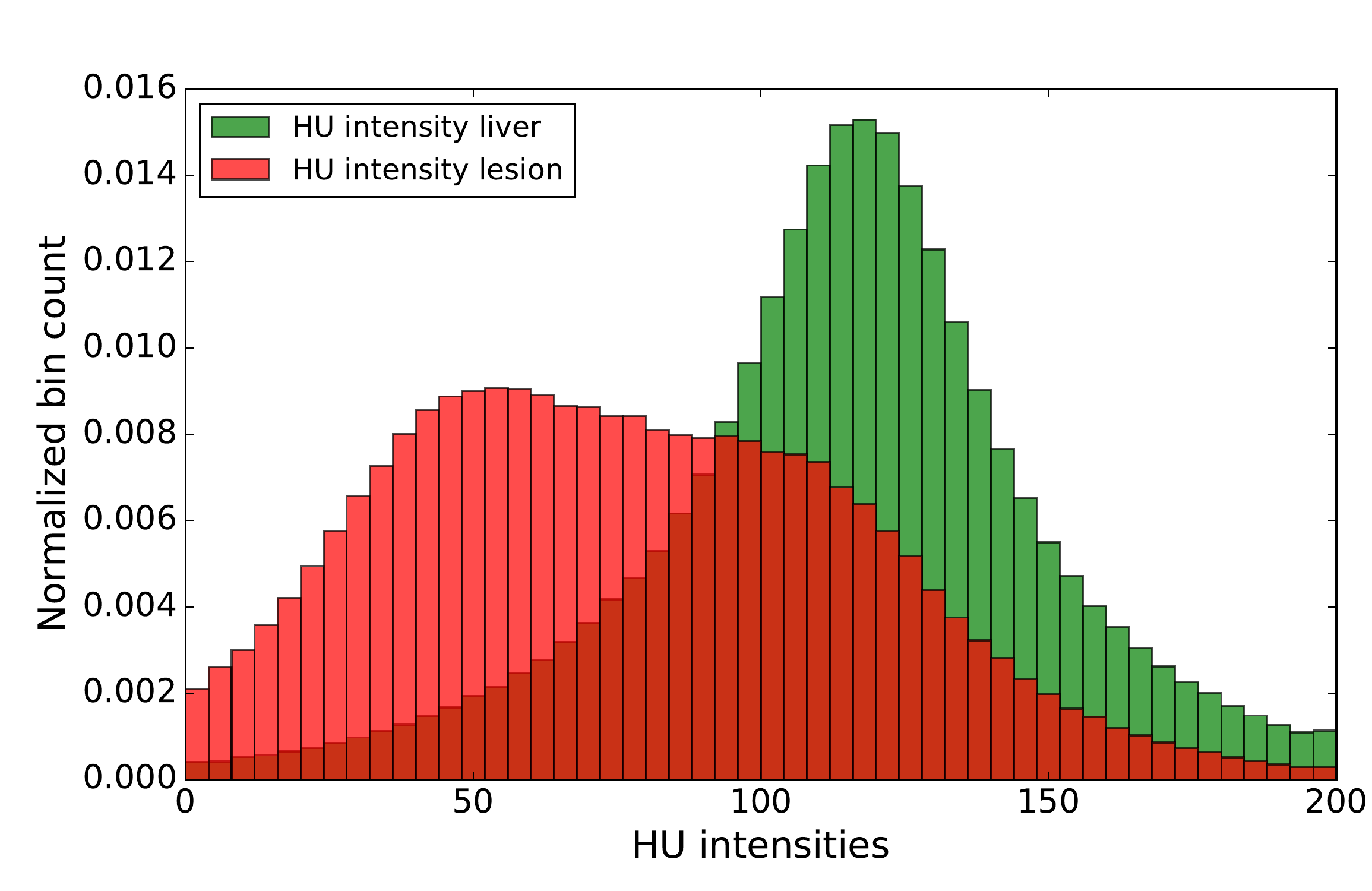}

\endminipage\hfill
\minipage[c]{0.32\textwidth}
\centering
        \includegraphics[width=\textwidth]{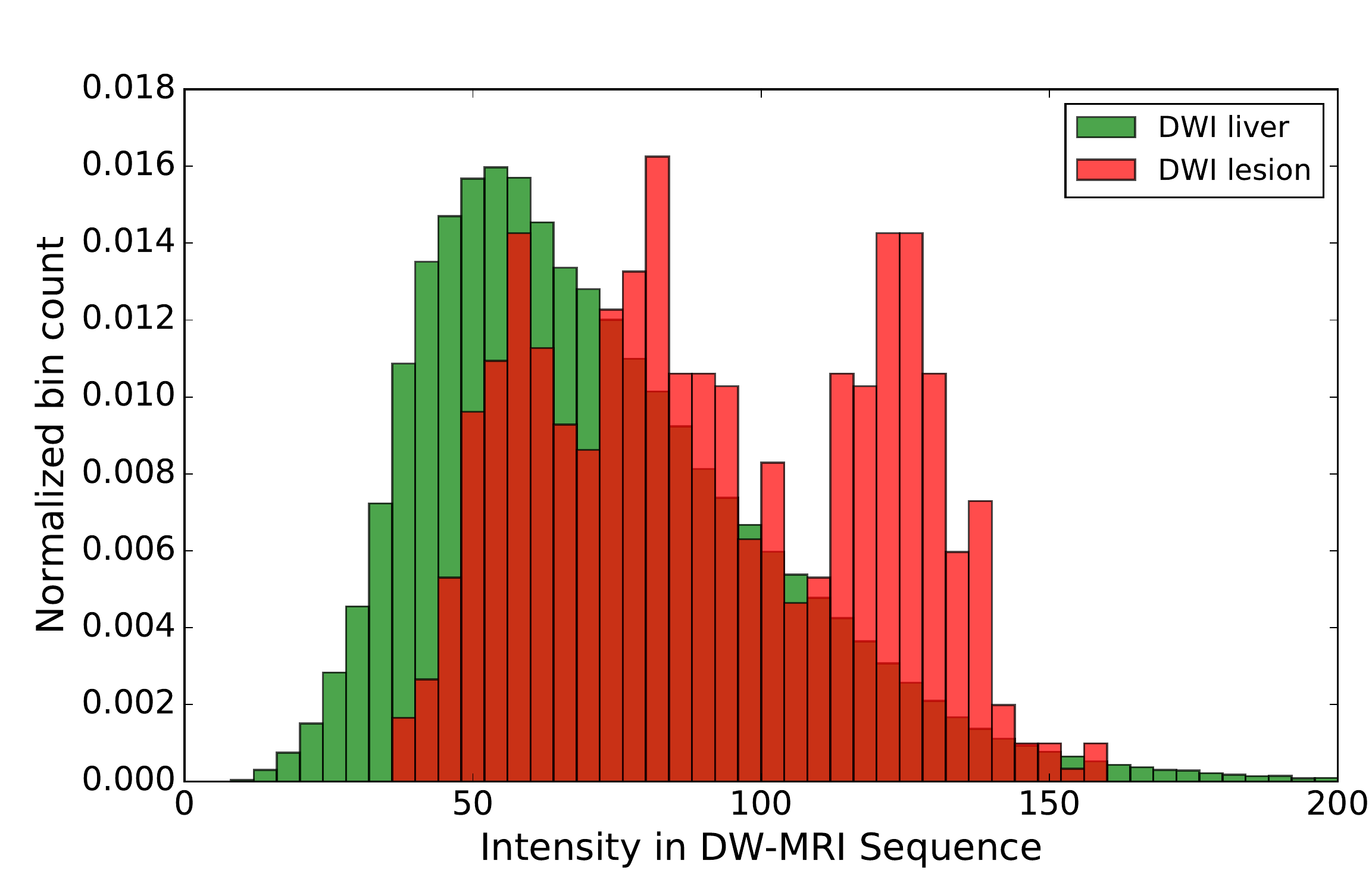}

\endminipage\hfill

    \caption{Liver and liver lesions slices in CT and diffusion weighted DW-MRI as well as the corresponding histogram for liver and lesions pixels in the respective modality. The shape, size and level of contrast vary for different lesions. As the histograms indicate, there is a significant overlap between liver and lesion intensities, leading to a low overall contrast.}
    \label{fig:motivation}
\end{figure}

In clinical routine, manual or semi-manual segmentation techniques are applied to interprete CT and MRI images that have been acquired in the diagnosis of the liver. These techniques, however, are subjective, operator-dependent and very time-consuming. In order to improve the productivity of radiologists, computer-aided methods have been developed in the past. However, an automated robust segmentation of combined liver and lesion remains still an open problem because of challenges as a low-contrast between liver and lesion, different types of contrast levels (hyper-/hypo-intense tumors), abnormalities in tissues (such as after surgical resection of metastasis), size and varying number of lesions.
As shown in figure \ref{fig:motivation} the heterogeneity in liver and lesion contrast is very large among subjects. Different acquisition protocols, differing contrast-agents, varying levels of contrast enhancements and dissimilar scanner resolutions lead to unpredictable intensity differences between liver and lesion tissue. This complexity of contrast differences make it difficult for intensity-based methods to generalize to unseen test cases from different clinical sites. In addition, the varying shape of lesions due to irregular tumor growth and response to treatment (i.e surgical resection) reduce efficiency of computational methods that make use of prior knowledge on lesion shape.

\subsection{Related Works}

Nevertheless, several interactive and automatic methods have been developed to segment the liver and liver lesions in CT volumes. In 2007 and 2008, two Grand Challenges benchmarks on liver and liver lesion segmentation have been conducted in conjunction with MICCAI conference \cite{Heimann,deng2008editorial}. Methods presented at the challenges were mostly based on statistical shape models. Furthermore, grey level and texture based methods have been developed \cite{Heimann}. Recent work on liver and lesion segmentation employs graph cut and level set techniques \cite{li2015automatic,li2013likelihood,linguraru2012tumor}, sigmoid edge modeling \cite{foruzan2015improved} or manifold and machine learning \cite{kadoury2015metastatic,freiman2011liver,vivanti2015automatic,ben2015automated}. However, these methods are not widely applied in clinics, due to their speed and robustness on heterogeneous, low-contrast real-life CT data. To overcome these weaknesses, interactive methods were still developed \cite{hame2012semi} to overcome these weaknesses.

Deep Convolutional Neural Networks (CNN) have gained significant attention in the scientific community for solving computer vision tasks such as object recognition, classification and segmentation \cite{krizhevsky2012imagenet,long2014fully}, often out-competing state-of-the art methods. Most importantly, CNN methods have proven to be highly robust to varying image appearance, which motivates us to apply them to fully automatic liver and lesions segmentation in CT volumes. 

Semantic image segmentation methods based on fully convolutional neural networks FCN were developed in \cite{long2014fully}, with impressive results in natural image segmentation competitions \cite{chen2014semantic,zheng2015conditional}. Likewise, new segmentation methods based on CNN and FCNs were developed for medical image analysis, with highly competitive results compared to state-of-the-art. \cite{Unet,Brats,wang2015detection,prasoon13,kamnitsas2016efficient,roth2015deeporgan,chen2016voxresnet,stollenga2015parallel}. 

\subsection{Contribution}

In this work, we demonstrate the combined automatic segmentation of the liver and its lesions in low-contrast heterogeneous medical volumes. Our contributions are three-fold. First, we train and apply fully convolutional CNN on CT volumes of the liver for the first time, demonstrating the adaptability to challenging segmentation of hepatic liver lesions. Second, we propose to use a cascaded fully convolutional neural network (CFCN) on CT slices, which segments liver and lesions sequentially, leading to significantly higher segmentation quality, as demonstrated on a public challenge dataset. Third, we experimentally demonstrate the generalization and scalability of our methods to different modalities and diverse real-life datasets, including a novel diffusion weighted MRI dataset and a large multri-centric CT dataset.\\
A preliminary version of this work was presented in MICCAI 2016 \cite{Christ2016} and will be presented at ISBI 2017 \cite{isbi2017}. In this paper, we have substantially revised and extended these previous publications. The main modifications include an elaborated description of the proposed methods, an analysis of underlying design principles and architectures as well as the application to new datasets and modalities.
\\
\\
In the following sections, we will describe our proposed pipeline (\ref{sec:pipeline}) including CFCN (\ref{nn}) and 3D CRF (\ref{sec:3dcrf}). The experiments are illustrated in section (\ref{sec:results}).

\section{Methods}

\subsection{Overview of our Proposed Segmentation Workflow}
\label{sec:pipeline}
Our proposed segmentation workflow is depicted in figure \ref{pipeline}. The workflow consists of three major steps. The first step (e.g. section \ref{preprocessing}) deals with data preprocessing and preparation for the neural network segmentation. In a second step (e.g. section \ref{nn}) two cascaded fully convolutional neural networks first segment the liver and then lesions within the liver region-of-interest (ROI). In the final third step, the calculated probabilities of CFCN will be refined using a dense 3D conditional random field to produce the final segmentation result.
\begin{figure}
\resizebox{\textwidth}{!}{%
    \centering
    \documentclass[../main.tex]{subfiles}

\begin{tikzpicture}
  [node distance=.8cm,
  start chain=going below,]

      \node (preprocess) [punktchain ]  {\textbf{Preprocessing}\\ \small{CT: HU Windowing \\ MRI: N4 Bias correction}};
      \begin{scope}[start branch=venstre,
        every join/.style={->, thick, shorten <=1pt}, ]
        \node[punktchain, on chain=going left, join=by {<-}]
            (risiko) {\textbf{Training Data}\\ \small{CT/MRI Volumes}};
      \end{scope}
      \begin{scope}[start branch=hoejre,]
      \node (finans) [punktchain, on chain=going right ,join=by {<-}] {\textbf{Test Data}\\ \small{CT/MRI Volumes}};
      
    \end{scope}

  \node[punktchain, join,] (cfcn) {\textbf{FCN for Liver Segmentation}\\ \small{A first FCN segments liver from abdomen CT/MRI}};
      \begin{scope}[start branch=left,
                every join/.style={->, thick, shorten <=1pt}, ]
                \node[punktchain, on chain=going left, join=by {<-}] (var) {\textbf{Data Augmentation} \\ \small{Mirror, crop, addition of noise and elastic deformation}};
        \end{scope}

  \node[punktchain, join] (roi) {\textbf{Segmented Liver ROI}};
  \node[punktchain, join] (roi) {\textbf{FCN for Lesion Segmentation}\\ \small{A second FCN segments lesion from from segmented liver ROIs}};
      \begin{scope}[start branch=left,
                every join/.style={->, thick, shorten <=1pt}, ]
                \node[punktchain, on chain=going left, join=by {<-}] (var) {\textbf{Data Augmentation} \\ \small{Mirror, crop, addition of noise and elastic deformation}};
        \end{scope}

  \node[punktchain, join] (konk) {\textbf{Postprocessing}\\ \small{3D Conditional Random Field}};
    \node[punktchain, join] (konk) {\textbf{Final Segmented CT/MRI Volume}};

 
  \end{tikzpicture}
    }
    \caption{Overview of the proposed image segmentation workflow for training and testing. As the first step the CT/MRI volumes are preprocessed with either HU-windowing or N4 bias correction. During the training phase the training data is augmented to foster the learning of invariance against noise and deformations in medical data. The CT/MRI volumes are trained after pre-processing and data augmentation in a cascaded fully convolutional neural network (CFCN). A first FCN segments the liver from abdomen CT/MRI scans. This segmented liver region of interest ROI is the input for a second FCN, that segments lesions from the given segmented liver ROI. To gain the final segmented volume is refined afterwards using a 3D conditional random field 3D CRF.}
    \label{pipeline}
\end{figure}
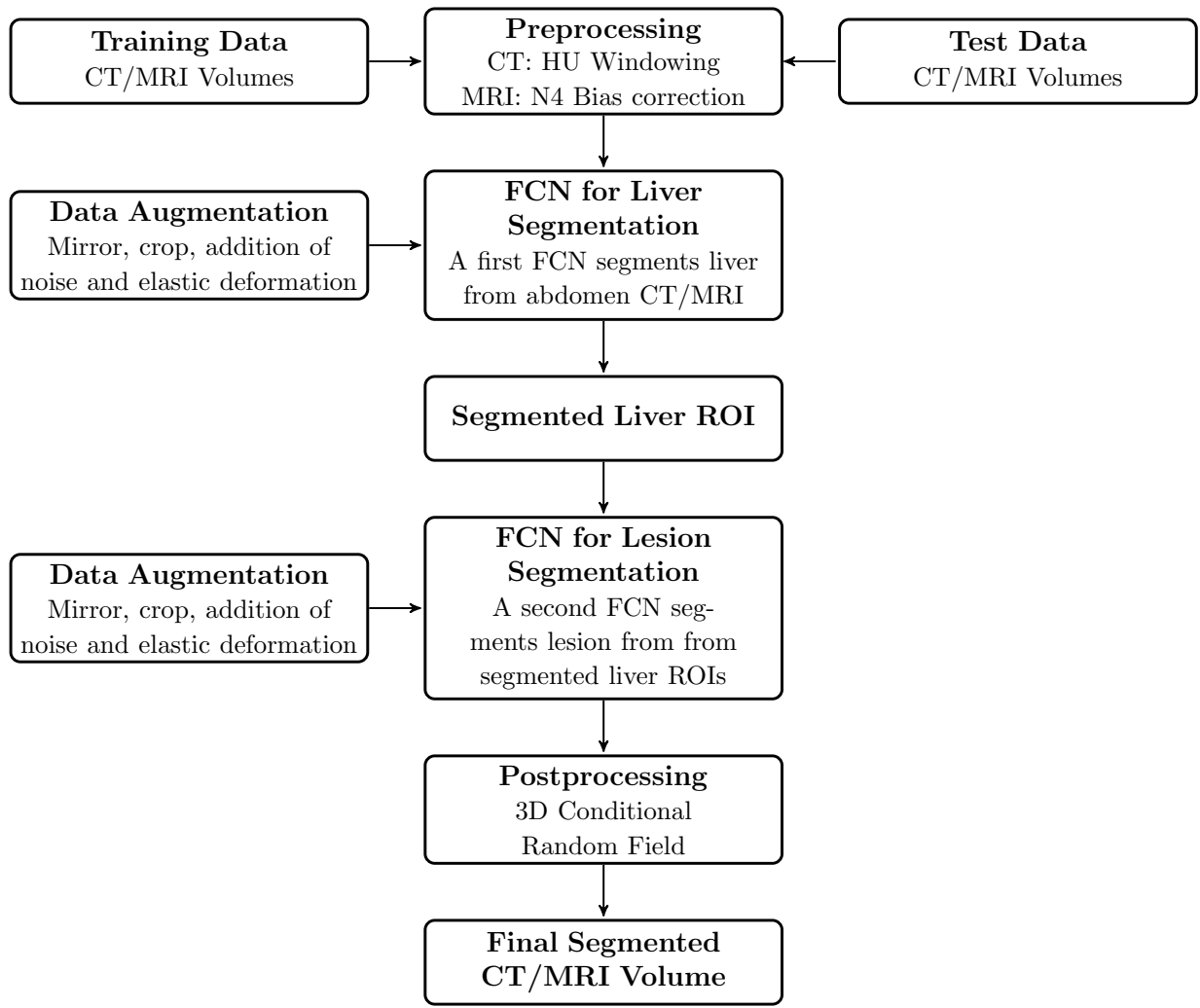

\subsection{Data Preparation}
\label{preprocessing}
The following section deals with data preprocessing and augmentation for CT data. 
Pre-processing was carried out in a slice-wise fashion. First, the Hounsfield unit values were windowed in the range $[-100,400]$ to exclude irrelevant organs and objects. Figure \ref{fig:preprocessing} shows the effect of our applied preprocessing to a raw medical slice. We increased contrast through histogram equalization. Figure \ref{fig:preprocessing} shows also the final slice after HU-windowing and contrast-enhancement. The contrast within the liver has been enhanced to allow better differentiation of abnormal liver tissue. For DW-MRI the data preparation scheme is similar and differs in the data normalization, which additionally performs N4bias correction \cite{Tustison2010}.

As in \cite{Unet,kamnitsas2016efficient}, to teach the network the desired invariance properties, several data augmentations steps, such as elastic deformation, translation, rotation and addition of Gaussian noise with standard deviation of the current slice, have been employed to increase the training data for the CFCN. Details on the data augmentation schemes is made available in our sourcecode\footnote{Sourcecode and models are available at https://github.com/IBBM/Cascaded-FCN}.
\begin{figure}[tb]

\minipage{0.32\textwidth}
    \includegraphics[width=\linewidth]{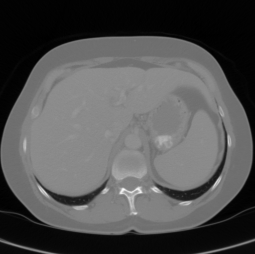}
\endminipage\hfill
\minipage{0.32\textwidth}
  \includegraphics[width=\linewidth]{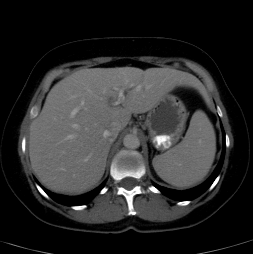}
\endminipage\hfill
\minipage{0.32\textwidth}%
  \includegraphics[width=\linewidth]{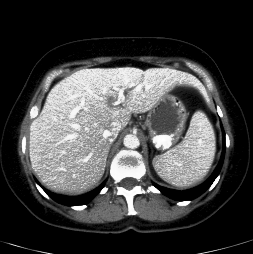}
\endminipage
\caption{Overview of the applied preprocessing steps. The raw CT slices (left) are windowed to a Hounsfield Unit range of -100 to 400 HU to neglect organs and tissues that are not of interest. The HU-windowed slice (middle) is further processed using a histogram equalization to allow further contrast enhancement of abnormal tissue (right). }
\label{fig:preprocessing}
\end{figure}

\subsection{Cascaded Fully Convolutional Neural Networks}
\label{nn}
In the following section, we describe different state-of-the the art deep learning architecture and design choices that we evaluated for a use in our segmentation tasks. We denote the 3D image volume as $I$, the total number of voxels as $N$ and the set of possible labels as $\labelspace  = \{0,1,\ldots,l\}$. For each voxel $i$, we define a variable $x_i \in \labelspace$ that denotes the assigned label. The probability of a voxel $i$ belonging to label $k$ given the image $I$ is described by $P(x_i=k \vert I)$ and will be modelled by the FCN. 
In our particular study, we use $\labelspace = \{0,1,2\}$ for background, liver and lesion, respectively.
\subsubsection{From AlexNet to U-Net}
Long et al. (2015) presented the first fully convolutional network architecture for semantic segmentation \cite{long2014fully}. The main idea in their work is to replace the last fully connected layers of a classification network such as the AlexNet \cite{krizhevsky2012imagenet} with fully convolutional layers to allow dense pixel-wise predictions. The last fully convolutional layers have to be upscaled to match the input dimensions. In comparison to prior work, the AlexFCN allows pixel-wise prediction from full-sized medical slices, instead of patch-wise classification. Figures \ref{fig:alexfcnliver} and \ref{fig:alexfcnlesion} show the training curves for training the AlexFCN (without class balancing) on 3DIRCAD dataset. Both training curves converged fast to a steady state in training and test Dice overlap. Both training curves show a large overfitting of the AlexFCN without class balancing, with Dice overlaps of 71\%/90\% in test/training data for liver, and 24\%/60\% for lesions. In general the lesion Dice of 24\% at test time is comparable low.
Long et al. (2015) explicitly stated that they did not need to apply class balancing to their natural image segmentation problem. A reason for this is that they used pretrained AlexNet weights trained on natural images, i.e. ImageNet data. However, for many medical applications it is mandatory to apply class balancing since pre-trained networks from natural images cannot be used properly and the class of interest occurs more seldomly in the dataset. Figures \ref{fig:liverclass} and \ref{fig:lesionclass} show the importance of class balancing in medical image segmentation. The training and test Dice for both liver and lesions increases noticeably to 78\% for liver and 38\% for lesions. 
A further large improvement can be obtained by applying the U-Net Architecture proposed by Ronneberger et al. (2015) \cite{Unet}. Besides the increased depth of 19 layers and learnable upscaling (up-convolution), the U-Net provides a superior design pattern of skip connections between different stages of the neural network. 

In early stages of the neural network, spatial information is present in the activations of the current stage. In later stages of the neural network, spatial information gets transferred to semantic information at the cost of specific knowledge on the localization of these structures. Here, for example, the original U-Net architecture reduces an input image of size 388x388 to a size of 28x28 in the U-Net bottleneck. Ronneberger et al. introduced skip-connections to allow utilization of spatial and semantic information at later stage, since the spatial information from earlier stage can be fused in the neural network at later stages. Thus the neural network at later stages can utilize semantic and spatial information to infer information.

\begin{figure}
\begin{subfigure}{.5\textwidth}
  \centering\captionsetup{width=.9\textwidth}
  \includegraphics[width=\linewidth]{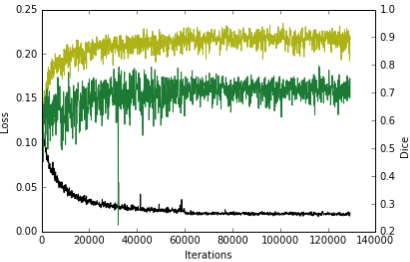}
  \caption{AlexFCN architecture without class balancing: Loss (black), Training Dice (light green), Test Dice (dark green) of Liver}
  \label{fig:alexfcnliver}
\end{subfigure}%
\begin{subfigure}{.5\textwidth}
  \centering\captionsetup{width=.9\textwidth}
  \includegraphics[width=\linewidth]{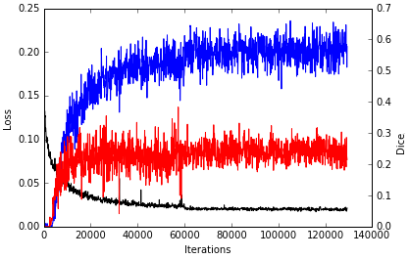}
  \caption{AlexFCN architecture without class balancing: Loss (black), Training Dice (blue), Test Dice (red) of Lesion}
  \label{fig:alexfcnlesion}
\end{subfigure}
\begin{subfigure}{0.5\textwidth}
  \centering\captionsetup{width=.9\textwidth}
  \includegraphics[width=\linewidth]{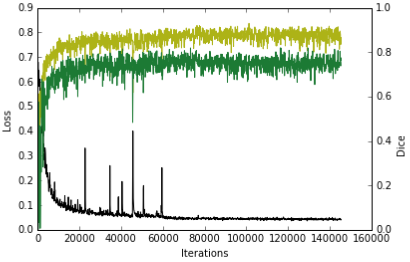}
  \caption{AlexFCN architecture with class balancing: Loss (black), Training Dice (light green), Test Dice (dark green) of Liver}
  \label{fig:liverclass}
\end{subfigure}
\begin{subfigure}{0.5\textwidth}
  \centering\captionsetup{width=.9\textwidth}
  \includegraphics[width=\linewidth]{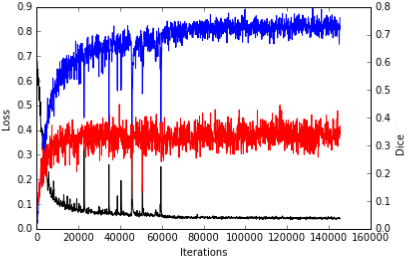}
  \caption{AlexFCN architecture with class balancing: Loss (black), Training Dice (blue), Test Dice (red) of Lesion}
  \label{fig:lesionclass}
\end{subfigure}
\begin{subfigure}{0.5\textwidth}
  \centering\captionsetup{width=.9\textwidth}
  \includegraphics[width=\linewidth]{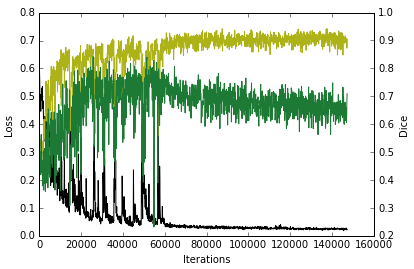}
  \caption{U-Net architecture with class balancing: Loss (black), Training Dice (light green), Test Dice (dark green) of Liver}
  \label{fig:unetliver}
\end{subfigure}
\begin{subfigure}{0.5\textwidth}
  \centering\captionsetup{width=.9\textwidth}
  \includegraphics[width=\linewidth]{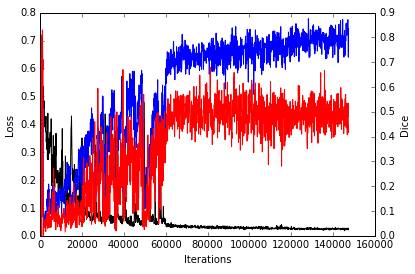}
  \caption{U-Net architecture with class balancing: Loss (black), Training Dice (blue), Test Dice (red) of Lesion}
  \label{fig:unetlesion}
\end{subfigure}
\caption{Training curves of different network architectures and training procedures of liver and lesion on 3DIRCAD dataset.}
\label{fig:fig}
\end{figure}

\subsubsection{From FCN to CFCN}
We used the U-Net architecture \cite{Unet} to compute the soft label probability maps $P(x_i \vert I)$. The U-Net architecture enables accurate pixel-wise prediction by combining spatial and contextual information in a network architecture comprising 19 convolutional layers. Figures \ref{fig:unetliver} and \ref{fig:unetlesion} show the training curves for the U-Net on 3DIRCAD data set. The overall performance of the lesion segmentation is further increased to 53\% test Dice. The U-Net learned features to discriminate liver and lesion at the same time.
As one of our main contributions, we propose a cascaded training of FCNs to learn specific features for solving a segmentation task once per training, which leads to higher segmentation performance.

The motivation behind the cascade approach is that it has been shown that U-Nets and other forms of CNNs learn a hierarchical representation of the provided data. The stacked layers of convolutional filters are tailored towards the desired classification in a data-driven manner, as opposed to designing hand-crafted features for separation of different tissue types. By cascading two U-Nets, we ensure that the U-Net in step 1 learns filters that are specific for the detection and segmentation of the liver from an overall abdominal CT scan, while the U-Net in step 2 arranges a set of filters for separation of lesions from the liver tissue. Furthermore, the liver ROI helps in reducing false positives for lesions.
Figures \ref{fig:cascade1} and \ref{fig:cascade2} illustrate our proposed method. We train one network to segment the liver in abdomen slices (step 1). This network can solely concentrate on learning discriminative features for liver vs. background segmentation, e.g. figure \ref{fig:cascade1}. After that we train another network to segment the lesions, given an image of the liver (step 2). The segmented liver from step 1 is cropped and re-sampled to the required input size for the cascaded U-Net in step 2. All non-liver regions are masked out and the second U-Net can concentrate on learning discriminative features for lesion vs. liver background segmentation.

\begin{figure}
    \centering
    \includegraphics[width=\textwidth]{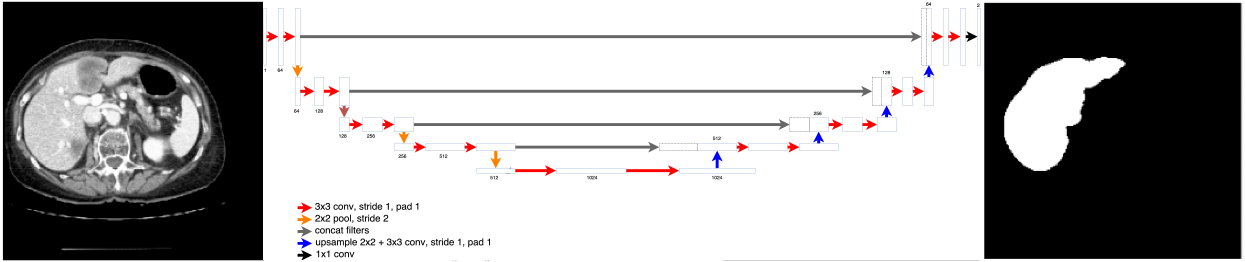}
    \caption{Step 1 of Cascaded FCN: The first U-Net learns to segment livers from a CT slice.}
    \label{fig:cascade1}
\end{figure}
\begin{figure}
    \centering
    \includegraphics[width=\textwidth]{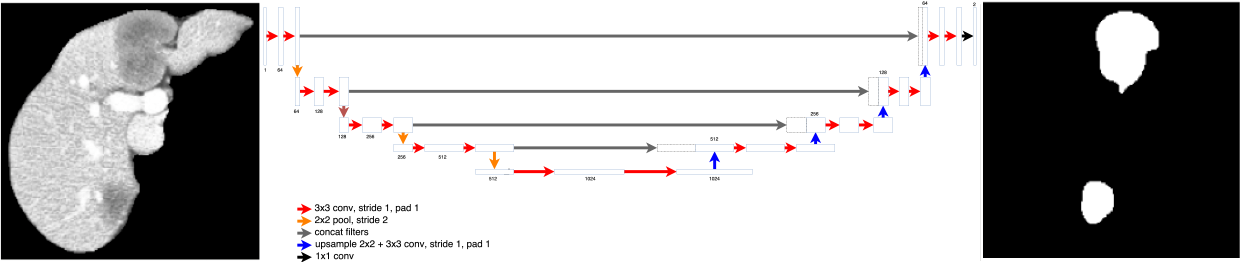}
    \caption{Step 2 of Cascaded FCN: The second U-Net learns to segment lesions from a liver segmentation mask segmented in step 1 of the cascade}
    \label{fig:cascade2}
\end{figure}
\subsubsection{Effect of Class Balancing}
\label{classbalancing}
A crucial step in training FCNs is appropriate class balancing according to the pixel-wise frequency of each class in the data. In contrast to \cite{long2014fully}, we observed that training the network to segment small structures such as lesions is not possible without class balancing, due to the high class imbalance that is typically in the range of <1\% for lesion pixels. Therefore we introduced an additional weighting factor $\omega^{class}$ in the cross entropy loss function $L$ of the FCN:
\begin{equation}
    L =- \frac{1}{n} \sum\limits_{i=1}^N \omega_i^{class}  \left[ \hat{P_i} \log P_i + (1 - \hat{P_i}) \log(1 - P_i) \right]
\end{equation}
$P_i$ denotes the probability of voxel $i$ belonging to the foreground, $\hat{P_i}$ represents the ground truth. We chose $\omega^{class}_i$ to be $\frac{\sum_i 1 - \hat{P_i}}{\sum_i \hat{P_i}}$ if $\hat{P_i} = 1$ and $1$ otherwise.

\subsubsection{Transfer Learning and Pretraining}
A common concept in deep learning is transfer learning using pretrained neural network models. Neural networks pretrained on a other task, e.g. a natural image classification data set, can be used as initialization of the network weights when training on a new task e.g. image segmentation of medical volumes. The intuition behind this idea is, that also for other tasks or dataset the first layers of neural networks learn similar concepts to recognize basic structures such as blobs and edges. This concepts do not have be trained again from scratch when using pretrained models. For our experiments we used pretrained U-Net models provided by Ronneberger et al. (2015), which were trained on cell image segmentation data \cite{Unet}. We have released our trained models on liver and lesion segmentation to allow other researcher to start their training with learned liver and lesion concepts\footnote{Sourcecode and models are available at https://github.com/IBBM/Cascaded-FCN}.
\subsection{3D Conditional Random Field}
\label{sec:3dcrf}
Volumetric FCN implementation with 3D convolutions was strongly limited by GPU hardware and available VRAM \cite{prasoon13}. Recent work such as V-Net and 3D U-Net, allow nowadays 3D FCNs at decreased resolution \cite{milletari2016v,3dunet}. In addition, the anisotropic resolution of medical volumes (e.g. 0.57-0.8mm in axial and 1.25-4mm in sagital/coronal voxel dimension in 3DIRCADb) complicates the training of discriminative 3D filters. Instead, to capitalise on the locality information across slices within the dataset, we utilize 3D dense conditional random fields (CRFs) as proposed by \cite{Krahenbuhl2012}. To account for 3D information, we consider all slice-wise predictions of the FCN together in the CRF applied to the entire volume at once.

We formulate the final label assignment given the soft predictions (probability maps) from the FCN as \emph{maximum a posteriori} (MAP)
inference in a dense CRF, allowing us to consider both
spatial coherence and appearance.

We specify the dense CRF following \cite{Krahenbuhl2012} on the
complete graph $\graph=(\vertices, \edges)$ with vertices $i \in
\vertices$ for each voxel in the image and edges $e_{ij} \in \edges =
\lbrace (i, j) \enspace \forall i, j \in \vertices \enspace
\mathrm{s.t.} \enspace i < j \rbrace$ between \emph{all} vertices. The
variable vector $\x \in \labelspace^N$ describes the label of each
vertex $i \in \vertices$. The energy function that induces the according
Gibbs distribution 
is then given as:

\begin{equation}
  \label{eq:energy}
  E(\x) = \sum_{i \in \vertices} \phi_i(x_i) + \sum_{(i,j) \in \edges} \phi_{ij}(x_i, x_j) \enspace 
\end{equation}
where $\phi_i(x_i) = -\log \cp{x_i}{I}$ are the unary potentials that
are derived from the FCNs probabilistic output,
$\cp{x_i}{I}$. $\phi_{ij}(x_i,x_j)$ are the pairwise potentials, which
we set to:
\begin{eqnarray}
  \label{eq:pairwisepot}
  \phi_{ij}(x_i, x_j) =  &\mu(x_i, x_j) \bigg( w_{\mathrm{pos}}  \exp  \left( -\frac{\vert p_i - p_j \vert^2}{2 \sigma_{\mathrm{pos}}^2} \right) \qquad \notag \\&+ w_{\mathrm{bil}} \exp \left( -\frac{\vert p_i - p_j \vert^2}{2 \sigma_{\mathrm{bil}}^2} -\frac{\vert I_i - I_j \vert^2}{2 \sigma_{\mathrm{int}}^2}\right) \bigg) \enspace 
\end{eqnarray}
where $\mu(x_i,x_j) = \mathbf{1}(x_i \neq x_j)$ is the Potts function,
$\vert p_i - p_j \vert$ is the spatial distance between voxels $i$ and
$j$ and $\vert I_i - I_j \vert$ is their intensity difference in the
original image. The influence of the pairwise terms can be adjusted
with their weights $w_{\mathrm{pos}}$ and $w_{\mathrm{bil}}$ and their
effective range is tuned with the kernel widths
$\sigma_{\mathrm{pos}}, \sigma_{\mathrm{bil}}$ and
$\sigma_{\mathrm{int}}$.

We estimate the best labelling $\x^* = \argmin_{\x \in \labelspace^N}
E(\x)$ using the efficient mean field approximation algorithm of \cite{Krahenbuhl2012}. The weights and kernels of the CRF were chosen using a random search algorithm adapted on the trainind data set.

\subsection{Quality Measures}
We assessed the performance of our proposed method using the quality metrics introduced in the grand challenges for liver and lesion segmentation by \cite{Heimann,deng2008editorial}.

Our main metric is the Dice score. Additionally we report Volume Overlap Error (VOE), Relative Volume Difference (RVD), Average Symmetric Surface Distance (ASD) and Symmetric Maximum Surface Distance (MSD). Metrics are applied to binary valued volumes, so a metric computed on the lesions for example considers only lesion objects as foreground and everything else as background. We refer to the foreground object in the ground truth as object A, and object B for the predicted object.
\subsubsection{Dice Score (DICE)}
The Dice score or F1 measure is evaluates as:
$$DICE(A,B) = \frac{2| A \cap B |}{|A| + |B|}$$
where the Dice score is in the interval [0,1]. A perfect segmentation yields a Dice score of 1.

\subsubsection{Volume Overlap Error (VOE)}

VOE is the just the complement of the Jaccard coefficient: 

$$ VOE(A,B) = 1 - \frac{| A \cap B |}{|A \cup B|}$$

\subsubsection{Relative Volume Difference (RVD)}

RVD is an asymmetric metric. It is defined as follows:

$$ RVD(A,B) = \frac{|B| - |A|}{|A|} $$

\subsubsection{Average Symmetric Surface Distance (ASD)}
Let $S(A)$ denote the set of surface voxels of $A$. The shortest
        distance of an arbitrary voxel $v$ to $S(A)$ is defined as:
        $$
            d(v,S(A)) = \min_{s_A\in S(A)} ||v-s_A||
        $$
        where $||.||$ denotes the Euclidean distance. The average symmetric
        surface distance is then given by:
$$
            ASD(A,B) =
                \frac{1}{|S(A)|+|S(B)|}
                \left(
                    \sum_{s_A\in S(A)} d(s_A,S(B))
                    +
                    \sum_{s_B\in S(B)} d(s_B,S(A))
                \right)
$$

\subsubsection{Maximum Surface Distance (MSD)}

MSD is also known as the Symmetric Hausdorff Distance. Maximum Surface Distance (MSD) is similar to ASD, except that the maximum distance is taken instead of the average.

$$
MSD(A,B) = \max
                \left\{
                    \max_{s_A\in S(A)} d(s_A,S(B)),
                    \max_{s_B\in S(B)} d(s_B,S(A)),
                \right\}
$$

\section{Experiments and Results}
\label{sec:results}
\label{dataset}
For clinical routine usage, methods and algorithms have to be developed, trained and evaluated on heterogeneous real-life data. In this work we want to demonstrate the robustness, generalization and scalability of our proposed method by applying it to a public dataset for comparison (section \ref{3dircad}), a clinical CT dataset (section \ref{clinicalct}) and finally a clinical MRI dataset (section \ref{clinicalmr}).

\subsection{3DIRCAD}
\label{3dircad}
\subsubsection{Dataset}

We evaluated our proposed method on the 3DIRCADb dataset\footnote{The dataset is available at \url{http://ircad.fr/research/3d-ircadb-01}} \cite{soler20123d}. In comparison to  the grand challenge datasets, the 3DIRCADb dataset offers a higher variety and complexity of livers and its lesions and is publicly available. The 3DIRCADb dataset includes 20 venous phase enhanced CT volumes from various European hospitals with different CT scanners. For our study, we trained and evaluated our models using the 15 volumes containing hepatic tumors in the liver with 2-fold cross validation. The analyzed CT volumes differ substantially in the level of contrast-enhancement, size and number of tumor lesions (1 to 42).
\subsubsection{Experimental Setting}
Data was prepared as described in section \ref{preprocessing}. Our data augmentation scheme lead to a total training data size of 22693 image slices. The CFCN were trained on a recent desktop PC with a single NVIDIA Titan X GPU with 12 GB VRAM. The neural networks were implemented and trained using the deep learning framework caffe \cite{jia2014caffe} from University of Berkeley. We used stochastic gradient descent as optimizer with a learning rate of 0.001 and a momentum of 0.8. To reduce overfitting we applied a weight decay of 0.0005.
\subsubsection{Effect of Class Balancing}
The effect of class balancing can be seen in figure \ref{fig:alexfcnliver} - \ref{fig:lesionclass}. Introducing class balancing improved the segmentation Dice score on both liver and lesion, while simultaneously decreasing over-fitting. The effect is less for liver, since the percentage of liver voxels in a CT abdomen dataset is on the order of 7\%, in comparison to 0.25\% for lesions. For all following experiments we accounted for class imbalance by weighting the imbalanced class according to its frequency in the dataset by introducing a weight factor described in section \ref{classbalancing}.
\subsubsection{Qualitative and Quantitative Results}
The qualitative results of the automatic segmentation are presented in figure \ref{fig:3dircad}. The complex and heterogeneous structure of the liver and all lesions were detected in the shown images. The cascaded FCN approach yielded an enhancement for lesions with respect to segmentation accuracy compared to a single FCN as can be seen in figure \ref{fig:3dircad}.
In general, we observe significant\footnote{Two-sided paired t-test with p-value $< 4 \cdot 10^{-19}$ } additional improvements for Dice overlaps of liver segmentations, from mean Dice $93.1\%$ to $94.3\%$ after applying the 3D CRF. 
For lesions we could achieve a Dice score of $56\%$ at a standard deviation of $26\%$ with a 2 fold cross-validation. 
\begin{figure}
    \centering
    \includegraphics[width=\textwidth]{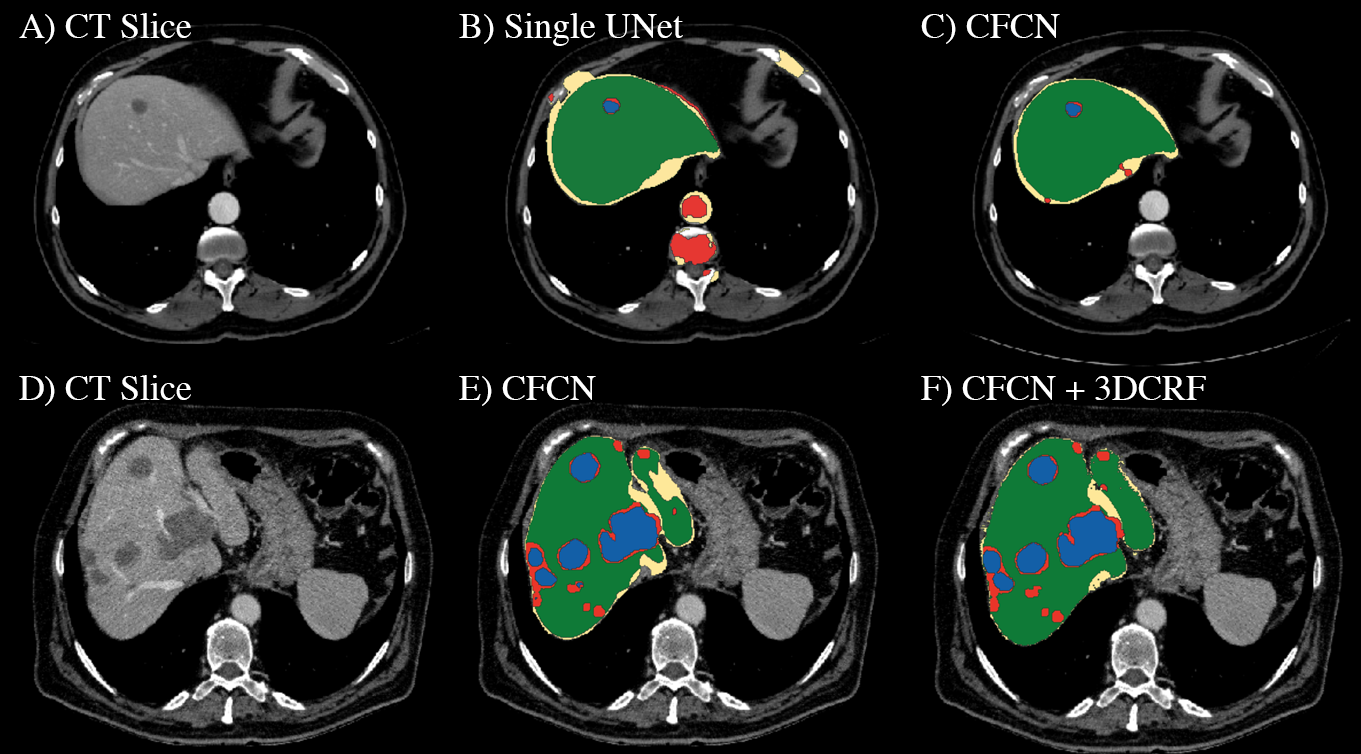}
    \caption{Automatic liver and lesion segmentation with cascaded fully convolutional networks (CFCN) and dense conditional random fields (CRF). Green depicts correctly predicted liver segmentation, yellow for liver false negative and false positive pixels (all wrong predictions), blue shows correctly predicted lesion segmentation and red lesion false negative and false positive pixels (all wrong predictions). In the first row, the false positive lesion prediction in B of a single U-Net as proposed by \cite{Unet} were eliminated in C by CFCN as a result of restricting lesion segmentation to the liver ROI region. In the second row, applying the 3D CRF to CFCN in F increases both liver and lesion segmentation accuracy further, resulting in a lesion Dice score of 82.3\%.}
    \label{fig:3dircad}
\end{figure}

\begin{table}[]
\centering
\resizebox{\textwidth}{!}{%
\begin{tabular}{@{}llcccccc@{}}
\toprule
Approach& Dataset & VOE & RVD & ASD & MSD & DICE \\
   & & {[}\%{]} & {[}\%{]} & {[}mm{]} & {[}mm{]} & {[}\%{]} \\ \midrule
 U-Net as in \cite{Unet} &3DIRCAD & $39$ & $87$  & $19.4$ & $119$  & 72.9 \\
 Cascaded U-Net &3DIRCAD  & 12.8  & -3.3 & 2.3 & 46.7 &  93.1\\
 Cascaded U-Net + 3D CRF&3DIRCAD  & 10.7 & -1.4 & 1.5 & 24.0  & 94.3 \\
  &  &  &  &  &  &  \\
 Li et al. \cite{li2015automatic} (liver-only)&3DIRCAD  & $9.2$ & $-11.2$ & $1.6$ & $28.2$&  \\
 Chartrand et al.  \cite{chartrand2014semi} (semi-automatic) &3DIRCAD & $6.8$ & $1.7$ & $1.6$ & $24$&  \\
Li et al.  \cite{li2013likelihood} (liver-only)&3DIRCAD  &  &  &  & & 94.5  \\ \midrule
 Cohen et al. \cite{ben2016fully} (liver-only)&Own Clinical CT  &  &  &  &   & 89 \\
 Cascaded U-Net&MR-DWI  & 23 & 14 & 5.2 & 135.3  & 87 \\
  Cascaded U-Net&Clinical CT  & 22 &  -3 & 9.5 & 165.7 & 88 \\
    Cascaded U-Net + 3D CRF&Clinical CT  & 16 & -6 & 5.3 & 48.3  & 91 \\
  \bottomrule
\end{tabular}%
}
\caption{Quantitative segmentation results of the liver on the 3DIRCADb dataset and other clinical CT am MR-DWI datasets. Scores are reported as presented in the original papers.}
\label{tab:benchmarking}
\end{table}

\subsection{Clinical Dataset CT}
\label{clinicalct}
\subsubsection{Dataset}

The second dataset we evaluated is a real-life clinical CT dataset from multiple CT scanners and acquired at different centers. It compromises 100 CT scans from different patients. The examined patients were suffering from different kind of cancerous diseases with different manifestations in the liver. The dataset ranges from single HCC lesions to diffusive and confluent metastatic lesions. In addition different contrast agents and therefore different levels of contrast enhancement are present in this dataset. Human rater ground truth was obtained through manual volumetric segmentation using the software TurtleSeg\footnote{www.turtleseg.com} \cite{top10turtleseg,top11turtleseg}.
\subsubsection{Experimental Setting}
The clinical CT dataset was prepared and augmented in the same way as the 3DIRCAD dataset as described in \ref{preprocessing}. The data set was split in 60 for training, 20 for test and 20 for validation. The neural networks, where trained on the same setup and training parameters as the 3DIRCAD dataset. In this experiment, an Adam optimizer was applied with $\epsilon=0.1$ \cite{kingma2014adam}.

\subsubsection{Qualitative and Quantitative Results}
As shown in table \ref{tab:benchmarking} the Cascaded FCN and Cascaded FCN + 3D CRF reach up to 88\% and 91\% Dice score on this dataset. An inter-rater Dice comparison among 5 training cases yielded a Dice overlap score of 95\%. Considering the inter-rater Dice score, the proposed method provides remarkable segmentations. 
Furthermore, our proposed method achieves a Dice overlap score of $61\% \pm 25\%$ for lesions on the validation set.

\subsection{Clinical Dataset MRI}
\label{clinicalmr}
\subsubsection{Dataset}
To demonstrate the generalization to other modalities we employed our methods to a clinical DW-MRI dataset. 31 Patients underwent clinical assessment and MR imaging for the primary diagnosis of HCC. Imaging was performed using a 1.5 T clinical MRI scanner (Avanto, Siemens) with a standard imaging protocol including axial and coronal T2w, axial T1w images before and after application of Gadolinium-DTPA contrast agent. Diffusion weighted imaging was performed using a slice thickness of 5mm and a matrix size of 192 by 192. The human rater ground truth segmentation was created for the DW-MRI sequence to allow further automatic image analysis e.g. section \ref{survival}.

\subsubsection{Experimental Setting}
In comparison to the CT datasets, the DW-MRI dataset was prepared differently. The DW-MRI dataset was normalized using the N4Bias correction algorithm \cite{Tustison2010}. Afterwards the same pre-processing steps were carried out as for CT. The CFCN for the DW-MRI dataset, where trained on the same hardware and training setup. The optimizer in this experiment was an Adam optimizer with $\epsilon=0.1$.

\subsubsection{Qualitative and Quantitative Results}
As seen in figure \ref{fig:mr}, the CFCN was able to segment the liver lesion correctly. In both cases the CFCN undersegments the lesion leading to a Dice score of 85\% in both cases.
The quantitative segmentation results are shown in table \ref{tab:benchmarking}. The Cascaded U-Net was able to reach a dice score for liver in MR-DWI of 87\%. For lesion we found a mean dice score of 69.7\%.

\begin{figure}
\begin{subfigure}{.5\textwidth}
  \centering
  \includegraphics[width=\textwidth]{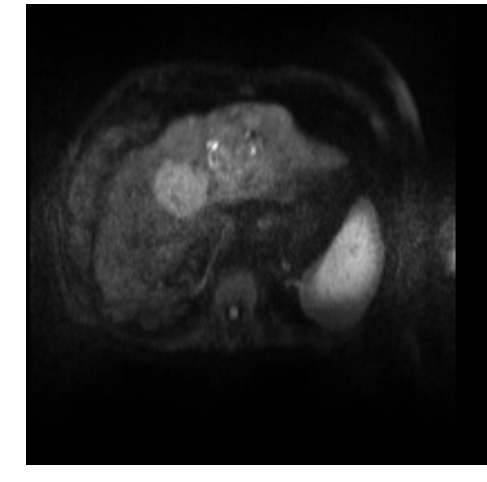}

\end{subfigure}%
\begin{subfigure}{.5\textwidth}
  \centering
  \includegraphics[width=\textwidth]{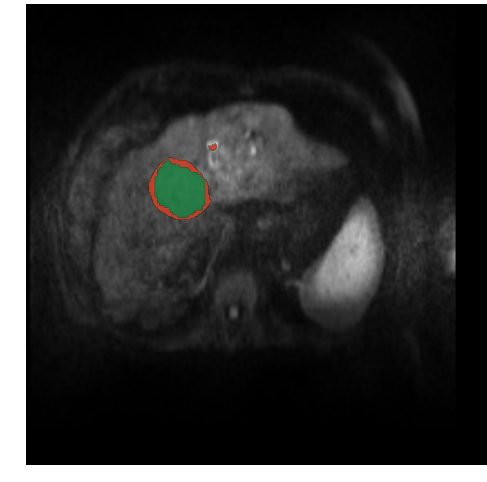}

\end{subfigure}

\begin{subfigure}{0.5\textwidth}
  \centering
  \includegraphics[width=\textwidth]{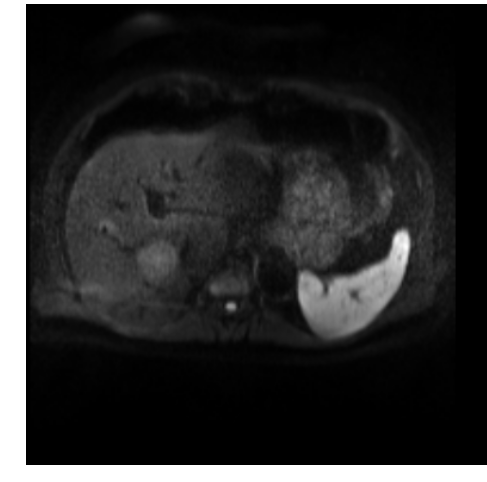}

\end{subfigure}
\begin{subfigure}{0.5\textwidth}
  \centering
  \includegraphics[width=\textwidth]{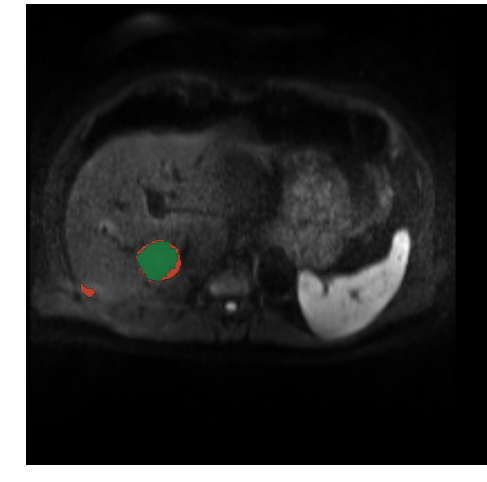}

\end{subfigure}
\caption{Automatic lesion segmentation with cascaded fully convolutional neural networks (CFCN) in DW-MRI. The raw DW-MRI slices (left), were automatically segmented with our proposed method. Green depicts correctly segmented lesion pixels. Red shows false positive and false negative, i.e. all wrong predictions, of the lesions. In both cases the proposed CFCN achieves an dice score for lesions of 85\%.}
\label{fig:mr}
\end{figure}

\subsection{HCC Survival Prediction Based on Automatic Liver and Lesion Segmentation}
Accurate liver and lesion segmentation are necessary for advanced medical image analysis and are meant to be input to radiomics algorithms, such as the SurvivalNet predictor \cite{isbi2017}. In this paragraph we want to introduce a possible applications of our automatic liver and lesion segmentation algorithms in medical imaging.
\label{survival}
Survival and outcome prediction are important fields in medical image analysis. For hepatic- cellular carcinoma HCC, prior work relied on manual liver and lesion segmentation in DW-MRI to calculate features over the liver and lesion ROI in the ADC sequence to predict patient survival. In contrast to prior work, we trained a CFCN to automatically segment liver and lesion segmentation in DW-MRI to allow automatic survival predictions. We formulate this task as a classification problem with classes being ``low risk" and ``high risk" represented by longer or shorter survival times than the median survival.
We predict HCC malignancy in two steps: As the first step we automatically segment HCC tumor lesions using our proposed method of cascaded fully convolutional neural networks (CFCN). As the second step we predict the HCC lesions' malignancy from the HCC tumor segmentation in the MR-DWI sequence using classical texture features and 3D CNN features. As one of our main contributions we found, that the accuracy of end-to-end assessment of tumor malignancy based on our proposed cascaded fully convolutional neural networks (CFCN) is equal to assessment based on expert annotations with high significance ($p>0.95$). In other words, our automatic tumor malignancy framework performs equally as assessment based on expert annotations in terms of accuracy.  Detailed information can be found in Christ, Ettlinger \& Kaissis et al. (2017) \cite{isbi2017}.  
\section{Discussion}
\subsection{Combined Segmentation and Clinical Relevance}
In comparison to state-of-the-art, such as \cite{foruzan2015improved,li2013likelihood,li2015automatic,chartrand2014semi}, we presented a framework, which is capable of a combined segmentation of the liver and its lesion. Moreover, we presented the clinical relevance of our proposed method by utilisation of our automatic segmentations to derive quantitative medical insights.
Furthermore, and in contrast to prior work such as \cite{Heimann,goryawala20123,lopez,Peng201481}, our proposed method could be generalized to segment the liver and lesion in different modalities and also multiple organs in medical data. As recent results from natural image segmentation indicate, fully convolutional networks are capable of segmenting dozens of labels with ease. By cascading the FCN architecture to smaller subregions the segmentation accuracy could be further increased. In addition with a runtime per slice of 0.19ms and 0.59ms our proposed method enables automatic segmentation of large-scale clinical trials in days and not months  \footnote{Estimating 3000 CT volumes for a large-scale clinical trial} using a single desktop PC.
\subsection{3D CNN and FCN Architectures}
Recent works such as DeepMedic \cite{kamnitsas2016efficient}, the V-Net \cite{milletari2016v} and the 3D U-Net \cite{3dunet} became possible due to efficient implementations of 3D convolutions on GPUs, and they show promising results on their respective segmentation tasks. The proposed idea of cascaded FCN could also be applied to novel 3D CNN and 3D FCN architectures. The restriction of the Region of Interest ROI to relevant organs as shown for the 2D U-Net, when restricting to liver only pixels for segmenting lesions, significantly boosts the segmentation accuracy. The intuition that more specific filters for the underlying problem could be trained, when restricting the relevant regions, holds for 3D as well. Future work will show whether 3D architectures could cope with less training data available for lesion segmentation.
\subsection{3D Conditional Random Field}

We showed a statistically significant improvement of segmentation quality, when applying the 3D CRF to our segmentation problem. However, tuning of hyperparameters such as those of the 3D CRF is very time-consuming and task dependent. We found that for highly heterogeneous structures in shape and appearance, such as HCC lesions, it is hard to find a hyperparameter set that generalizes to unseen cases with a random search. A similar conclusion was made in \cite{kamnitsas2016efficient} when applying a 3D CRF to heterogeneous brain lesions. Recent work successfully integrated the learning of the CRF hyperparameter in the training process \cite{zheng2015conditional}. This approach in combination with additional pairwise terms that incorporate prior knowlegde of the problem could lead to a improvement of the CRF for this task.

\section{Conclusion}
Cascaded FCNs and dense 3D CRFs trained on CT volumes are suitable for automatic localization and combined volumetric segmentation of the liver and its lesions. Our proposed method competes with state-of-the-art. We provide our trained models under open-source license allowing fine-tuning for other medical applications in CT data \footnote{Trained models are available at \url{https://github.com/IBBM/Cascaded-FCN}}.
Additionally, we introduced and evaluated dense 3D CRF as a post-processing step for deep learning-based medical image analysis. Furthermore, and in contrast to prior work such as \cite{foruzan2015improved,li2013likelihood,li2015automatic}, our proposed method could be generalized to segment multiple organs in medical data using multiple cascaded FCNs. 
As future work, the application of further cascaded FCNs on lesions ROIs to classify malignancy of the lesions as well as advanced techniques such as data augmentation using adversarial networks could enhance the accuracy of the segmentation further.
All in all, heterogeneous CT and DW-MRI volumes from different scanners and protocols can be segmented in under 100s each with the proposed approach. We conclude that CFCNs are promising tools for automatic analysis of liver and its lesions in clinical routine and large-scale clinical trials.

\label{summary}
\section{Acknowledgement}
This work was supported by the German Research Foundation (DFG) within the SFB-Initiative 824 (collaborative research center), ``Imaging for Selection, Monitoring and Individualization of Cancer Therapies" (SFB824, project C6) and the BMBF project Softwarecampus. We thank NVIDIA and Amazon AWS for granting GPU and computation support. 


\bibliographystyle{elsarticle-num}

\bibliography{literature}

\end{document}